\documentclass{ieeeaccess}
\usepackage{cite}
\usepackage{amsmath,amssymb,amsfonts}
\usepackage{graphicx}
\usepackage{textcomp}
\def\BibTeX{{\rm B\kern-.05em{\sc i\kern-.025em b}\kern-.08em
    T\kern-.1667em\lower.7ex\hbox{E}\kern-.125emX}}

\usepackage{amsmath,amssymb,amsfonts}
\usepackage{graphicx}
\usepackage{textcomp}

\usepackage{url}
\usepackage{color}
\usepackage{multirow}
\usepackage{verbatim}
\usepackage{subfigure}
\usepackage{tabularx}
\usepackage{colortbl}
\usepackage{booktabs}
\usepackage{xspace}
\usepackage{times}
\usepackage{enumerate}
\usepackage{shortvrb}
\usepackage{algorithm,algorithmicx}
\usepackage{caption}
\usepackage{algpseudocode}
\usepackage{mathtools}
\usepackage[mathscr]{eucal}
\usepackage{times}
\usepackage[pdfborderstyle={/S/U/W 1}]{hyperref}

\usepackage[multiple]{footmisc}
\usepackage[all]{nowidow}
\usepackage{balance}
\usepackage{adjustbox}

\newtheorem{problem}{Problem}

\definecolor{yellowdef}{RGB}{255,255,255}

\newcommand{\modify}[1]{\textrm{\textcolor{black}{#1}}}

\newcommand{\eat}[1]{}

\newcommand{\Patk}[1]{\mbox{\Pat@$#1$}}
\newcommand{\RBPatp}[1]{\mbox{\RBP@$#1$}}

\newcommand{\NDCGatk}[1]{\mbox{\NDCG@$#1$}}
\newcommand{\ERRatk}[1]{\mbox{\ERR@$#1$}}

\newcommand{\gbrt}{\method{GBRT}}
\newcommand{\lstm}{\method{LSTM}}
\newcommand{\dqn}{\method{DQN}}
\newcommand{\dodqn}{\method{DoDQN}}
\newcommand{\doddqn}{\method{DoDDQN}}
\newcommand{\ddqn}{\method{DDQN}}

\newcommand{\pdoddqn}{\method{PER-DoDDQN}}
\newcommand{\per}{\method{PER}}

\newcommand{\xgboost}{\texttt{XGBoost}}

\newcommand{\tensorflow}{\texttt{tensorflow}}

\newcommand\method[1]{{\sf\small{#1}}}

\newcommand{\argmax}{\operatornamewithlimits{argmax}}

\newcommand{\var}[1]{\mbox{\emph{#1}}}
\newcommand{\svar}[1]{\mbox{\scriptsize\emph{#1}}}

\newcommand{\myurl}[1]{{\url{#1}}}
\newcommand{\mycaption}[1]{\caption{\normalfont{#1}}}

\newcommand{\myparagraph}[1]{\vspace{0.2\baselineskip}\noindent{\textbf{#1}}.~}

\newcommand{\mycomment}[1]{}

\newlength{\onedigit}
\newcounter{todocount}
\newcommand{\figvspacelen}{-3ex}

\begin{document}
\history{Date of publication xxxx 00, 0000, date of current version xxxx 00, 0000.}
\doi{10.1109/Access-2019-17918}

\title{Which Channel to Ask My Question?: Personalized Customer Service Request Stream Routing using Deep Reinforcement Learning}


\author{
	\uppercase{Zining Liu}\authorrefmark{1},
	\uppercase{Chong Long\authorrefmark{2}, Xiaolu Lu\authorrefmark{3}, Zehong Hu\authorrefmark{4}, Jie Zhang\authorrefmark{5}, Yafang Wang\authorrefmark{2}}}

\address[1]{School of Software, Shandong University, Jinan 250101, Shandong, China(e-mail: liuzining@mail.sdu.edu.cn)}
\address[2]{Ant Financial Services Group, Z Space No. 556 Xixi Road Hangzhou,  310099, Zhejiang, China(e-mail: \{huangxuan.lc;yafang.wyf\}@antfin.com)}
\address[3]{RMIT University, GPO Box 2476, Melbourne VIC 3001 Australia (e-mail: xiaolu.lu@rmit.edu.au )}
\address[4]{Alibaba Group, 969 West Wen Yi Road, Hangzhou 311121, Zhejiang, China(e-mail: HUZE0004@e.ntu.edu.sg )}
\address[5]{School of Computer Science and Engineering, Nanyang Technological University, 50 Nanyang Avenue, Singapore, 639798(e-mail: zhangj@ntu.edu.sg)}

\tfootnote{This project is sponsored by National Natural Science Foundation of China ( No .61503217)}

\markboth
{Author \headeretal: Preparation of Papers for IEEE TRANSACTIONS and JOURNALS}
{Author \headeretal: Preparation of Papers for IEEE TRANSACTIONS and JOURNALS}

\corresp{Corresponding author: Yafang Wang (e-mail: yafang.wyf@antfin.com).}

\begin{abstract}
Customer services are critical to all companies, as they may directly connect to the brand 
reputation.
Due to a great number of customers, e-commerce companies often employ multiple communication channels to answer 
customers' questions, for example, chatbot and hotline.
On one hand, each channel has limited capacity to respond to customers' requests, 
on the other hand, customers have different preferences over these channels.
The current production systems are mainly built based on business rules, which 
merely considers tradeoffs between resources and customers' satisfaction.
To achieve the optimal tradeoff between resources and customers' satisfaction,  
we propose a new framework based on deep reinforcement learning, which directly takes both 
resources and user model into account.
In addition to the framework, we also propose a new deep-reinforcement-learning  based routing method -- double dueling deep Q-learning with prioritized experience replay (\pdoddqn).
We evaluate our proposed framework and method using both synthetic and 
a real customer service log data from a large financial technology company.
We show that our proposed deep-reinforcement-learning based framework is superior to the 
existing production system.
Moreover, we also show our proposed {\pdoddqn} is better than all other deep Q-learning variants in
practice, which provides a more optimal routing plan.
These observations suggest that our proposed method can seek the trade-off where both channel 
resources and customers' satisfaction are optimal.
\end{abstract}

\titlepgskip=-15pt

\maketitle

\section{Introduction}
\label{sec:introduction}
The quality of customer service is crucial to a company's reputation: its quality 
is measured by how quickly a company responds to customers' requests and 
how satisfied customers are when seeking help.
Obviously, the satisfaction of  a customer can merely be measured using the 
problem-solving quality alone.
In practice, a company also adopts customers' queuing time as one 
of the indicators for measuring satisfaction.
For the purpose of shortening customers' waiting time, major companies often 
provided multiple communication channels for customers to choose, for 
example, mobile App, web-based message and the traditional hotline.
Different communication channels have their own limited quota for responding to customers' requests,  and also require different cognitive loads. Customers are impatient, especially when they have a request to be resolved, but they want to express their need using the least effort. 
The most natural communication channel is the hotline, where a customer 
service representative answers the call and help customers solve their 
problems. As a consequence, most customers prefer the hotline channel and it often leads to a long waiting time, especially during peak hours.
Therefore, the obvious problem occurred is the imbalanced workloads of each channel.
How can we solve this problem to make customers satisfied, and how can we help companies optimally allocate limited resources are the key questions.
There are a lot of other factors that impact customer acceptance rates, e.g., the design of user interface, the privacy issues of the question, etc. All of them can affect user's choice. This paper focus only on the routing problem at first: redistributing the right customers to non-hotline, and reducing the burden of a round of dialogue for unsuitable customers. In the experiment result section we can see the significant improvements after applying our routing model. We will consider other factors in our future work.


The customer service system produces about 50000 data every day in the e-commerce company. 
Existing systems adopted by many companies are rule-based to deal with big data issues, which are easy to implement as most of the business rules are already defined.
However, they are not flexible and often end up with a large set of complicated 
rules.
Also, using a rule-based system doesn't aim to seek a balance between optimality of resource allocation and customer satisfaction, but purely for the low-cost implementation purpose. We carefully explore and evaluate these drawbacks in this paper, and show that although rule-based systems may be an efficient way to handle customer requests stream, it is far from making both customers and the company satisfactory.


Motivated by the recent success in deep reinforcement learning~\cite{Sutton+Barto:2018}, we 
propose a deep-reinforcement-learning based framework to perform the 
customer service requests routing.
The proposed framework is in sharp contrast to the rule-based system,
as our framework directly captures: (i) customers' preference and (ii) 
each channels future traffic.
Customers have different preferences over different communication channels, 
and a simple routing method may result in low satisfaction.
For example, if the problem is not so urgent, a student may be willing to leave 
messages on the offline service desk if the current hotline is busy; someone 
who is doubtful about the chatbot may be stick to the hotline channel, 
regardless the waiting time. 
Recommending customers their preferrable  surrogate channel is essential as 
it affects the overall customers' experience.
Another key to solving the allocation problem is to be able to predict the 
channel's traffic over the  next time window.
The requests data stream often comes in a high speed, especially in peak 
hours  and a mass corporation, a system can fail miserably if it ignores the 
time-series feature of the request stream.
For example, if our prediction  suggests that the hotline is not busy in the next 
time window we can  then let customers be served using the hotline if they 
prefer to; similarly if our prediction shows the hotline's capacity will be 
exceeded in the next time window, we may try to route customers' request 
to alternative channels if possible.
Hence, the framework we proposed in this paper is  to seek tradeoffs among 
the channel capacity, user preference  and the predicted traffic of the current 
channel.

Our framework consists of three major components: (i) customer profiling module, (ii) flow forecasting module and (iii) an agent that suggests a customer to the most suitable channel.
The first component is used to model the {\em environment} in our reinforcement learning framework and the last one is the {\em agent}.
We make use of users' attribute to infer their preference over different channels, so that our system 
can perform personalized routing according to each individual's preference.
As we mentioned, the customer requests stream is a time-series data and it is crucial if we can 
predict the volume of each channel of a certain time point.
In this paper, we evaluate several techniques for time-series prediction and give a recommended 
approach.
Our framework is built based on deep reinforcement learning method.
Despite of using existed deep Q-learning and its variants,  we also propose the
\textbf{d}ouble \textbf{d}ueling {\dqn} with \textbf{p}rioritized \textbf{e}xperience 
\textbf{r}eplay (\per-\doddqn) method, which combines the 
strength from both double {\dqn} (\dodqn)~\cite{DBLP:conf/aaai/HasseltGS16} 
and dueling {\dqn} (\ddqn)~\cite{DBLP:conf/icml/WangSHHLF16}.

We evaluate our proposed  framework and method using  synthetic and real customer service data 
collected from a large financial technology company.
We propose several evaluation metrics and evaluate our solution effectiveness from three aspects: (i) 
channel congestion, (ii) customers' acceptance rate of the recommended alternatives and (iii) the channel utilization.
Experiments demonstrate the substantial routing effectiveness gains can be 
achieved -- both 
channel resources and customers satisfaction reach an optimal state -- using our 
proposed new routing framework is more effective than the existing system, and our proposed {\per-\doddqn} is more effective than {\dqn} variants.


Our contribution can be concluded in three-fold:
\vspace{-0.5ex}
\begin{enumerate}
\item  We  model the customer service requests routing problem using deep 
reinforcement learning, considering both channel resources and customers' 
satisfaction.
\item  We propose the double dueling deep Q-learning with prioritized 
experience 
replay method to solve the routing problem, which achieves that better 
performance than its counterparts in practice.
\item  We perform an extensive evaluation using both real and synthetic data to 
demonstrate the practical value of our proposed methods.
\end{enumerate}
\vspace{-0.5ex}
The rest of this paper is organized as follows. The background and our problem definition will be introduced in Section~\ref{sec-background}. After that, we will present our system, including the user model and the routing approach based on reinforcement learning	in Section~\ref{sec-system}, together with experimental results and analysis in Section~\ref{sec-experiments}. Then related technical work about reinforcement learning and traffic forecasting will be provided in Section~\ref{sec-related-work}. Finally, the conclusions and future work will be provided in Section~\ref{sec-conclusion}.

\section{Background}
\label{sec-background}
In this section
, we provide the background of routing application in detail, including application scenario and the baseline system.

\subsection{Application Background}
\label{subsec-app}
Modern business often provides several communication channels for customers convenience, 
ranging from  traditional call center service, online chatbot, to mobile APP.
Besides the traditional call center (or hotline) services, other channels may be served by a mixture of 
automatic chatbot and human customer representatives.
Apparently, the hotline service has the least capacity to deal with customers' request,  while it is the 
most preferred one by a majority of the customers, due to its low communication cost.
Other channels face a similar trade-off.
For example, it is easier for users to access mobile Apps for issuing  customer service requests,  but 
it requests a lot of cognitive effort for users to describe their information need precisely and the interaction 
is often less efficient compared to using the hotline service.
While most of the time the mobile APP may not be preferred, for simple requests, customers' may resolve 
them more efficient by selecting or browsing pre-selected questions.
Similarly, for the online interface, customers are required to input their requests precisely, which may  
prevent customers from using this channel. 
However, compared to the mobile APP and the hotline service, the web interface provides a way for 
customers to interact with each other, which  may help to reduce the workload of customer service 
representatives to some extent.
When a large number of customers are making requests to one channel, the channel's capacity may be 
exceeded, and customers need to wait a long time.
In this  case, most customer services platforms will re-direct users to other channels, however,  users have 
a large chance to reject such recommendation.

According to our application scenario, we need to propose a system that seeks tradeoffs between the 
channel capacity and customers' satisfaction, where the satisfaction is measured using the acceptance rate and 
waiting time.
Let $n$ be the total number of communication channels, each of which has a capacity of $c_i (1\leq i \leq 
n)$.
Assume all customers will use the customer service platforms.
Each customer has a user profile $\mathbf{u} = \langle{f_0, f_1,\dots, f_k }\rangle $, where $f_i$ represents a 
value of $i$-th attribute.
We consider at time $t$, each channel will also have a request flow, referred to as $e_t$.
The utilization rate $v_i$ to represent the percentage of the free resources of the current channel, relative
to the channel capacity.
Formally, the routing problem is:

\begin{problem}{\sc Customer Service Routing Problem.}
	\label{def-problem}
	Given a user and the request, and  the current capacity of each channel $c_i$, the problem is to 
	recommend 
	user a communication channel $i$, such that: (i) the overall users' satisfaction rate  is maximized; 
	(ii) the overall utilization rates of communication channels are maximized. 
\end{problem}

\subsection{Baseline Customer Service Routing System}
\label{subsec-rule-sys}

\begin{figure}
	\centering
\includegraphics[scale=0.5]{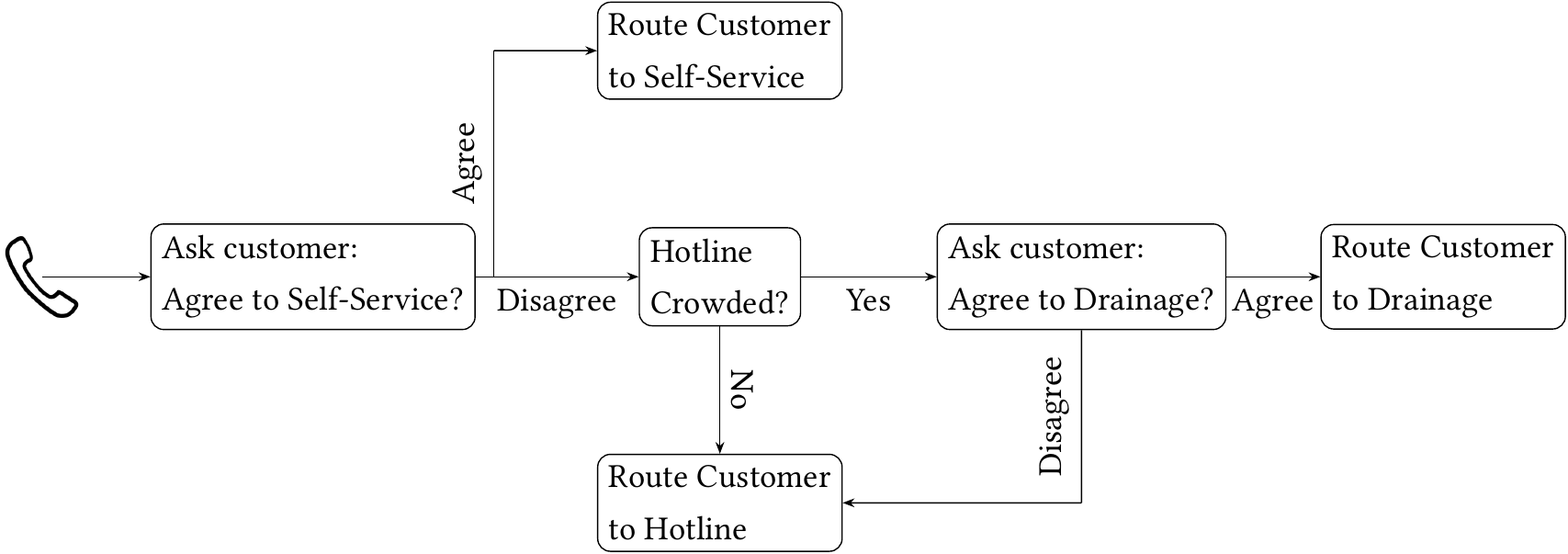}
\caption{Existing customer service request routing system, which is designed 
based on  business rules.}
\label{online-structure}
\vspace{\figvspacelen}
\end{figure}
\modify{To compared with our proposed RL-based algorithm,} we consider a real customer service routing system as an example, shown in 
Figure~\ref{online-structure}.
This is a real product system adopted by a large financial technology company, which is designed based
on business rules.
The system considers two communication channels: self-service and hotline, where self-service
let customers solve simple requests and the hotline is the traditional call-based customer service.

The system works as follows:
when a customer calls in, the system always asks whether the customer would like to use a 
self-service, as the hotline may be congested and the customer's question may be simple.
If the customer agrees, he or she will be routed to the self-service channel, otherwise, the system 
will then make a decision depending on the current queue length of the hotline channel.
A customer will always be re-directed to hotline service when there are no people waiting in the queue.
However, when the service channel is busy and there is a queue, the system will randomly select a 
set of customers to ask if they are willing to switch to the drainage channel.
The exact number of customers who will be chosen to ask for a second preference depends on the current 
length of the hotline queue -- if the queue is long, then more customers will be selected than a short 
queue.

From the description, we can observe that, existing rule-based customer service routing system 
strictly stick to the rule that always respects customers' choices, regardless of the current channel
state.
Customers can choose to stay wait until the hotline is available, or switch to other provided channels.
We analyze the real data collected from the production system, and find that only 20.1\% customers 
agreed to accept the self-service channel in the current customer service routing system.
A detailed analysis and comparison will be provided in Section~\ref{sec-experiments}.

Obviously, existing systems only provide a simple interface for users to express their interest, rather 
than performing an effective routing operation.
Letting customers choose their preferred channel almost always lead to the congestion in the 
hotline channel, as a result, customers will need to wait for a long time until be served; 
the other channels will be wasted in this situation.
In this paper, we want to mitigate this extremely imbalanced resources utilization, so we propose a 
new routing framework that aims at finding optimal plans for both the company and customers.

\section{Personalized Requests Routing Framework}
\label{sec-system}
As we described in Definition~\ref{def-problem}, the customer request routing 
system needs to find an optimal state between channel resources and 
customers'  satisfaction.
We first describe our proposed routing framework and then our proposed 
\textbf{d}ouble \textbf{d}ueling {\dqn} with \textbf{p}rioritized \textbf{e}xperience 
\textbf{r}eplay (\per-\doddqn) model.

\subsection{Framework Description}
\begin{figure*}[t]
	\centering
	\includegraphics[scale=0.8]{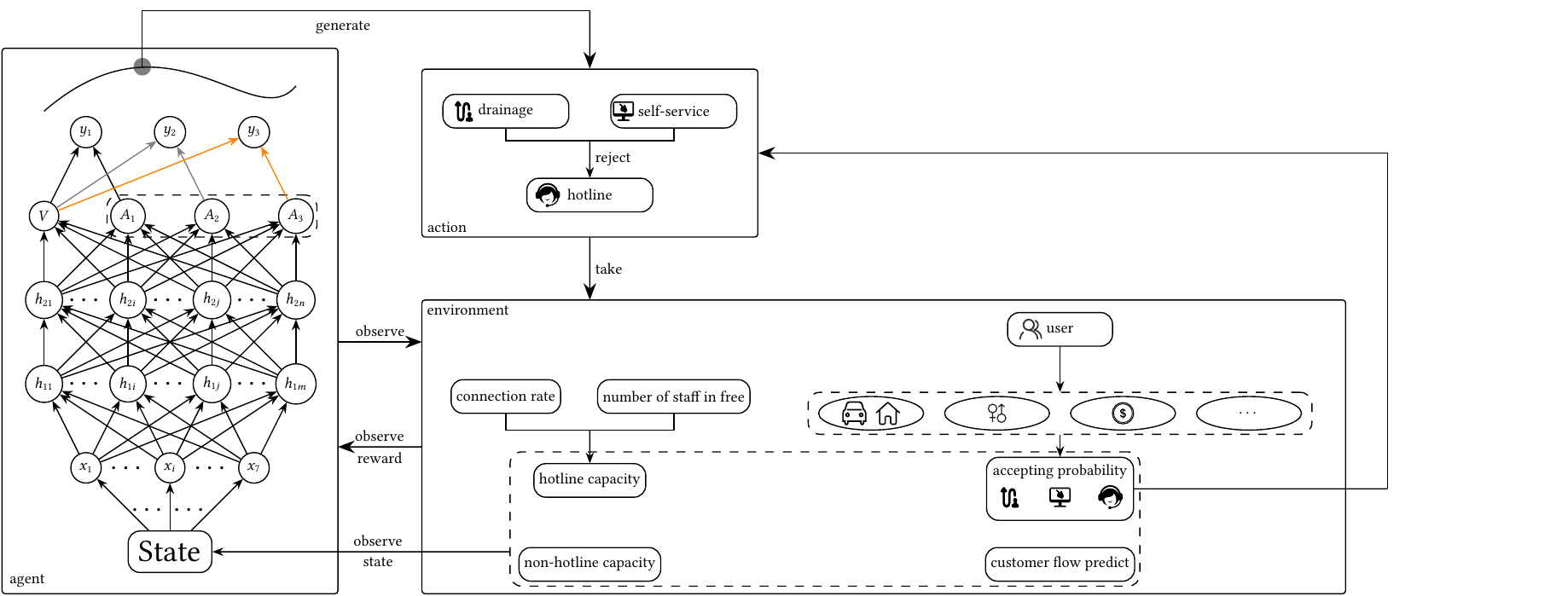}
	\mycaption{Overview of the proposed routing framework. 
	}\label{fig-architechture}
    \vspace{\figvspacelen}
\end{figure*}

We show an overview of  our proposed customer service routing system in 
Figure~\ref{fig-architechture},  which is based on deep reinforcement learning.
In the proposed framework, there are two important  components: environment 
and agent.

The environment is shown in the right bottom pane of Figure~\ref{fig-architechture},
which consists of a channel model (the left-hand side in the pane) and a user model 
(the right-hand side in the pane).
The channel model is used to estimate the volume of request flow, as a part of the 
state that the agent can observe.
If we estimate that a channel will have a large incoming request in the next time window, our routing framework should avoid allocating more customers to that channel.
Since the customer service request show a strong temporal correlation, we 
formulate the problem of channel requests predication as a time-series prediction 
problem.
Moreover, as in peak hours, the data stream may come in with a high volume and 
velocity, so pre-estimating the number of the incoming requests helps the system 
to make a better routing decision.
All flow forecasting algorithm described in Section~\ref{subsec-flow-forecasting} 
can potentially be applied to this problem.
After a careful evaluation, we choose the {\xgboost} as it reveals the best 
performance in the real scenario.
We will describe the detailed evaluation in Section~\ref{sec-experiments}.

\begin{figure}
    \centering
    \includegraphics[scale=1.0]{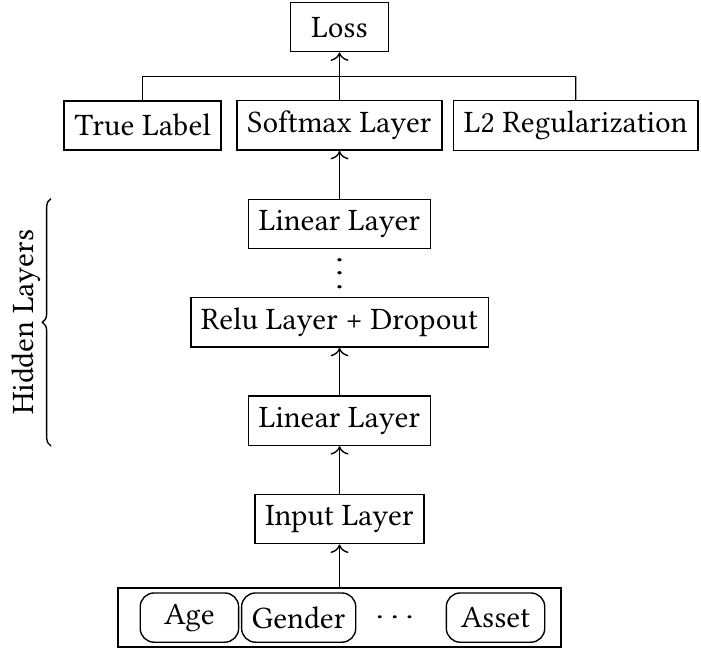}
    \caption{The user model.}
    \label{fig-userporfile}
\end{figure}

Another important element of the environment is the user model, describing  
customers' preference over different communication channels.
Intuitively, customers preferences vary on all possible communication channels.
If the recommended channel is not preferred  by  a customer, then it is highly likely 
the customer will reject the channel routing, which leads to a longer waiting queue 
of a popular channel, for example, the hotline channel.
We build the user model using a neural networks, 
which has three hidden layers with size 128, 256 and 128 respectively. 
For each user, we input their attributes into the user model and output their 
acceptance probability of each communication channel.
All routing models in this paper use the same architecture shown in Figure~\ref{fig-userporfile}.

The agent which takes actions based on  observed states is shown in the leftmost pane in 
Figure~\ref{fig-architechture}.
It takes the state observed from the environment as the input and then a neural 
network is used to recommend  a channel to the customer.
If a customer accepts our recommendation, then a successful routing is performed, 
which indicates we  may reduce the workload of some bottleneck channels to some 
extent and also the user is willing to accept a channel switch.
Note that, in our framework,  all components can be configured with the most 
suitable algorithm, and we will  show an empirical comparison among different 
configurations in Section~\ref{sec-experiments}.


\subsection{The Routing Model}
The overall routing model is based on deep reinforcement learning, and more 
specifically, the {\dqn} variants. \modify{They approximate $q_\theta(s,a)$, which means the value of "state"(s) and "action"(a) pairs with the deep learning. The algorithms optimize the policy according to the TD-error: $r + \gamma \max \limits_{a'} q(s',a') - q(s,a)$ and are updated as the following equation}

\begin{equation}
q(s,a) = q(s,a)+ \alpha(r + \gamma \max \limits_{a'} q(s',a') - q(s,a)).
\label{eqn-qval-0}
\end{equation}

\modify{Where $r$ represents the immediate reward, and $s'$ and $a'$ represents the next state and the next possible actions. All {\dqn} variants are based on the above ideas and will be described in detail in Section \ref{sec-related-work}.}

We start to describe how we formulate the three key elements (action, state and reward function) in any 
reinforcement learning model,  and then we describe our proposed {\per-\doddqn} 
in Algorithm~\ref{algortihmflow}.

\myparagraph{Action}
The system  aims to learn a policy that can determine which channel should be 
recommended to users, so the action here is to select a channel among $n$ 
candidates, where $n$ is determined by the total number of channels in the real 
application.
We represent the action as $a,\,(a \in \{1, 2,3, \dots,n\})$.

\myparagraph{State}
A state is expected to express the customer's channel preference, 
and channel feasibility of handling furthermore requests.
So the state includes: the customers' preferences over different channels 
$\mathbf{u}$, 
the channel capacity $\mathbf{c}$ and the channel's future request flow traffic $\hat{\mathbf{e}}_t$.
The entire state is represented as $\mathbf{s} = \langle{\mathbf{u}, 
\hat{\mathbf{e}}_t, \mathbf{c}}\rangle$.


\myparagraph{Reward Function}
There are duel reward aspects we need to consider: from the customers' 
perspective and from the channel capacity, based on which the reward is:
\begin{equation}
\label{reward_function}
  \begin{split}
  R=g_{a, t}&-\lambda_1 \cdot \var{ReLu}\left(-\min(\mathbf{c}_{t}-\lambda_3\cdot\hat{\mathbf{e}}_{t+1} )\right) \\
  &-\lambda_2\cdot (\var{ReLu}(\min(\mathbf{c}_{t} - \lambda_3\cdot\hat{\mathbf{e}}_{t+1} )))^2,
  \end{split}
\end{equation}
where $\lambda_1 \gg  \lambda_2$ and $\lambda_3$ is a smoothing parameter, 
making reward  put more emphasis on the current capacity.

We use $g_{a, t}$ be the reward that is given to the user's current action at
time $t$, indicating if we make a good recommendation from a customer's 
perspective.
When a user accepts the recommendation,  $g_{a, t}=1$; otherwise 
$g_{a, t} = -1$.
Therefore, $g_{a, t}$ can guide the system learning towards the channel 
that a customer prefers.

The second and third terms in Equation (~\ref{reward_function}) is designed 
according to bottleneck channels, which is the channel has the least remaining 
capacity and favored by the majority, for example, the hotline channel.
As the bottleneck channel may easily be congested and intuitively, customers'
satisfaction will also be largely affected by the waiting time in the queue -- even 
more important than the preference.
To guide the system learn to avoid such case, we also add an extra quadratic punishment item to our reward function.

\myparagraph{Personalized Customer Service Routing Algorithm}
We describe the work flow using our proposed {\per-\doddqn} in 
Algorithm~\ref{algortihmflow}. The system takes channel capacity, user profile and the  request flow series as input, output a channel recommendation for each customer. As our focus is neither customer profiling nor flow forecasting, we simplify the two module description in our presentation. Note that, the choice of algorithm used to solve the problem should be made according to the real application data.
\begin{algorithm}[t]
\caption{Personalized Customer Service Routing (\per-\doddqn)}
\begin{algorithmic}[1]
\State {\bf Input:} Customer service flow, Channel capacity $C\gets C_0=(c_1, c_2, \dots, 
c_n)$, User attributes, period K, \textit{done\_num}
\State Build $\mathbf{u}(u_1,u_2,\dots, u_n )$
 \Comment{User model, and bottleneck channel's probability is 1.}
\State $D \gets \{\,\}, \theta \gets 0, \bar{\theta}\gets 0$
\Comment{Initialize {\per} buffer $D$, and parameters}
\State $A^* \gets \{\}$, terminal\_state$\gets$ False \Comment{Keep tracking 
customers' actions}
\State $T \leftarrow length(training\,dataset)$
\For {$t \leftarrow 1$ \textbf{to} $T$}
\State $c_{\svar{id}} \gets$ current customer's id 
\Comment{Get current customers' id}
\State $\mathbf{s}_t \gets \langle\mathbf{u}, \hat{e}, C\rangle$, $a_t \gets -1$
\label{stp-state}
\If{$t  > 1$}
\State {store transition$(\mathbf{s}_{t-1},a_{t-1},R_{t},s_{t})$ in $\mathcal{D}$ with weight
	\label{stp-replay-store} 
and update $\mathcal{D}$~\cite{DBLP:journals/corr/SchaulQAS15}}
\EndIf
\State {select action $\mathbf{a}_t$ according to 
		$\epsilon$-greedy~\cite{DBLP:conf/icml/WunderLB10}, following policy 
		$\pi_{\theta}$~\cite{DBLP:conf/aaai/HasseltGS16}\cite{DBLP:conf/icml/WangSHHLF16}, let $g_{a,t} \gets 1$}
		\If {customer reject the selected action}
		\State {reassign the customer to bottleneck channel, $g_{a,t} \gets -1$.}
		\EndIf
		\State {$A^*[c_{\svar{id}}] \gets a_t$, $C[a_t] \gets C[a_t]  - 1$}
		\label{stp-update-channel-start}
		\Comment{Reduce channel capacity}
		\State {Compute the immediate reward $R_{t+1}$ using 
		Equation~\ref{reward_function}
		}
		
		\If { a customer $c_i$ finishes request}
		\State {$C[A^*[c_i]] \gets C[A^*[c_i]]+1 $}
		\EndIf
				\label{stp-update-channel-end}
		\If {$\min(C)<$ \textit{done\_num}}
        \State {$s_t$ is terminal state}
        \State {reset the channel capacity:$C\gets C_0$}
		\EndIf
		
		\State{For the transition $(\mathbf{s}_{t-1}, {a}_{t-1}, 
		{R}_{t},\mathbf{s}_{t})$ sampled from $\mathcal{D}$ according to~\cite{DBLP:journals/corr/SchaulQAS15}, compute loss	
		\label{stp-doddqn-start}:
			\begin{equation}
\mathcal{L} = \left\{
			\begin{array}{lr}
			(R_{t} -q_{\theta}(\mathbf{s}_{t-1},a_{t-1}))^2, 
		\mbox{if $\mathbf{s}_{t}$ is terminal state} &  \\
			(R_{t}+\gamma 
			q_{\bar{\theta}}(\mathbf{s}_{t},{a}_{t})-q_{\theta}( 
			{s}_{t-1},{a}_{t-1}))^2,
			\mbox{otherwise} &  
			\end{array}
			\right.
			\label{loss-eq}
			\end{equation}
		}
	    \State{Where $q_{\theta}$ is computed based on
	    Eq~\ref{dueling-q-value}}	\label{stp-doddqn-end}
        \State{Minimize loss by gradient descent}
        \State{$\bar{\theta}\gets\theta$ if $t\bmod K \equiv 0$}
		\EndFor
	\end{algorithmic}
\label{algortihmflow}
\end{algorithm}
For incoming request stream, we first obtain the customer id and current channel 
capacity at the moment.
We then initialize the state at the current time point in Line~\ref{stp-state}.
If this is not the starting point and we've already made some actions, we store our previous action and current state into the replay buffer $\mathcal{D}$, in Line~\ref{stp-replay-store}. We then select the current optimal action for the current customer and update the channel capacity from Line~\ref{stp-update-channel-start} to Line~\ref{stp-update-channel-end}. Note that, a customer can choose to reject and accept our suggestion, and we need to update $g_{a,t}$ accordingly.
If a customer rejects our suggestion, we always put them into the bottleneck 
channel, which definitely will be  accepted.
From Line~\ref{stp-doddqn-start} to Line~\ref{stp-doddqn-end}, we describe the 
loss function used in our proposed {\per-\doddqn} method, in which it combines 
both {\dqn} and {\ddqn}.
As we've mentioned, we adopt prioritized experience replay, so the  mini-batch 
sampled from replay buffer is based on weights  computed in 
Equation~\ref{eqn-w-pre}.
Note that  when the environment reaches the terminal state, the 
TD-error is liable to suddenly increase.
 Therefore, we use prioritized experience replay so that the transition with the terminal state can be trained more times, speeding up the model convergence.

\section{Experiments}
\label{sec-experiments}
We describe experimental evaluation in this section, including results on  both 
synthetic and real datasets.
\subsection{Experimental Setup} \label{ep-set}
 \myparagraph{Real-World Dataset} 
Our real data is sampled from June 15, 2018 to October 1, 2018 from customer 
service log provided by a big financial technology company. 
In total, there are  4898143 customer requests when congestion happens. 
For each customer, we consider seven features, which consists of  gender, age, 
residential province, household, car, assets and credit limit. These features are preprocessed, and mapped to integer values if the original feature is categorical. 

 \myparagraph{Synthetic Dataset}
	For improving the robustness of RL model, we simulate data based on the real-world 
	Dataset. 
	Firstly, we extract 10 days data from the real-world dataset, then we 
	simulate the length of these sessions sampled from the uniform distribution, 
	$U(1, 6)$ minutes.
	Then we synthesize the probability of customer with R package 
	\method{SynthPop}~\cite{nowok2016synthpop}, with the cart method. 
	Finally, we counted customer flow and added Gaussian noises to the flow data as 
	the customer flow prediction data.
	We get 487{,}354 simulator	 customer service data from 15 June 2018 to 24 
	June 2018. We use 50{,}000 data from simulator dataset as the testing dataset.

It must be noted that the immediate reward according to Equation~\ref{reward_function} is considered as the labels in supervised machine learning because there are no real labels in the synthetic dataset and real-world dataset.

\myparagraph{Parameter Settings}
As there is a high percentage of noises in the real dataset,  we train all models using the synthetic dataset, and apply the optimal parameter  settings to the real 
dataset.
 For each configuration and the dataset, we list the parameter settings in 
Table~\ref{tbl-syn-param}.

 We select these parameters of RL models when the parameters fulfill these two conditions: (i) the loss of agent converging and (ii) the reward increment when the action is selected by the agent.
These models converge  with about 9000 iterations in simulator datasets, so the model is learned after the  experience replay is updated 50 times.


\begin{table*}[t]
    \centering
	\setlength{\tabcolsep}{4pt}
\begin{tabular}{ccccccccccccc}
	\toprule
	\multirow{2}{*}{Parameter} & \multicolumn{4}{c}{\per-\doddqn} &  & 
	\multirow{2}{*}{\doddqn} &  & \multirow{2}{*}{\dqn} &  & 
	\multirow{2}{*}{\dodqn} 
	&  & \multirow{2}{*}{\ddqn} \\ \cmidrule{2-5}
	& full & e & u & noter &  &  &  &  &  &  &  &  \\ \midrule
	$\lambda_1$ & 0.900 & 0.800 & 0.500 & 0.900 &  & 0.900 &  & 0.800 &  & 0.900 
	&  & 0.900 
	\\
		$\lambda_2$ & 0.015 & 0.020 & 0.015 & 0.015 &  & 0.015 &  & 0.015 &  & 
		0.020 &  & 0.015 \\
		$\lambda_3$ & 0.300 & -- & 0.300 & 0.300 &  & 0.300 &  & 0.300 &  & 0.300 &  
		& 0.300 \\
$\gamma$ & 0.500 & 0.300 & 0.700 & 0.500 &  & 0.500 &  & 0.700 &  & 0.500 &  & 
0.500 \\ 
	\bottomrule
\end{tabular}

	\mycaption{Parameter values obtained on the training set. Note we set the 
		memory 
		size to 2{,}000, and the batch size is 320. 
		For all models, we early terminate the current episode when the hotline 
		channel has a
		congestion of 100 requests.
	}\label{tbl-syn-param}
\end{table*}


\myparagraph{Baselines and Configurations}
In order to empirically show the effectiveness of our system, we consider the 
following variants:
(i) deep Q-Learning model variants and (ii) state variants.
So, in addition to our proposed system, double dueling DQN with prioritized replay 
(\pdoddqn), we consider the 
nature DQN~(\dqn), the dueling DQN~(\ddqn), a plain double DQN model~(\dodqn) and the double dueling DQN~(\doddqn) without prioritized replay.
In order to show the impact of our channel model, customer flow prediction model and terminal state, we also show results of which the state is $\langle{\mathbf{u}, \mathbf{c}}\rangle$,
$\langle{\mathbf{c}, \mathbf{\hat{e}}}\rangle$ and removing the terminal state, 
which we use {\pdoddqn-e}, {\pdoddqn-u} and {\pdoddqn-noter} to represent, 
respectively. 
All of our algorithms are implemented using {\tensorflow}.

\myparagraph{Traditional Machine Learning Algorithms}
In this paper, we also provide experiments by comparing traditional machine learning algorithms (KNN\cite{DBLP:journals/corr/abs-1712-05929}, CNNs\cite{DBLP:journals/wc/MaoTFKAIM18,noauthor_image-text_nodate} and SVM\cite{DBLP:journals/corr/abs-1712-05929}). In this paper, the Neural Network consists of two convolutional layers and two fully connected layers. 
LSTM cannot be applied in this task as the single message($<\mathbf{u},\mathbf{c},\mathbf{\hat{e}}>$) has no sequence information.

\myparagraph{Evaluation Metrics}
To detailly evaluate these models in simulator dataset, We designed six metrics. \modify{Where N represents the total number of consultations in the test set in the next six metrics.}
 {\em Congestion Rate}~(CCR) means the percentage of data when congestion 
 happens;
 \begin{equation}
    CCR = \frac{\sum \limits_{i=1} \limits^N \vec{1}_{C<0}}{N},
    \label{CCR-def}
\end{equation}
where $C<0$ means that there are people waiting for service.\modify{$\vec{1}_{C<0}$ represents whether the congestion occurs when people calling. If there are customers in the waiting queue, $\vec{1}_{C<0} = 1$, otherwise, $\vec{1}_{C<0} = 0$.}

{\em Average Congestion Level}~(AC) means the average degree of congestion in 
simulator dataset;
\begin{equation}
    AC = \frac{\sum \limits_{i=1} \limits^N Relu(-\vec{\mathbf{C}})}{N},
    \label{AC-def}
\end{equation}
where $Relu(-\vec{\mathbf{C}})$, $\vec{\mathbf{C}} = \{C_i\}$ represents the length of the waiting queue of channel $i$.
\modify{$C_i < 0$ means there number of customers in the waiting queue of channel $i$, $Relu(-\vec{\mathbf{C}})$ means if there are customers in waiting queue,   $Relu(C_i) = abs(C_i)$ .$C_i > 0$ means there are {$C_i$} capacity in channels, so $Relu(-C_i) = 0$}

{\em Peak congestion}~(PC) means the minimum total capacity.
\begin{equation}
    PC = max(Relu(-\vec{\mathbf{C}})),
    \label{AC-def}
\end{equation}
\modify{$PC$ refects the most congested degree, also represents the maximum length of the waiting queue.}

And {\em AFR} means the average idle degree of the catering staff.
\begin{equation}
    AFR = \frac{\sum \limits_{i=1} \limits^N Relu(\vec{\mathbf{C}})}{N},
    \label{AC-def}
\end{equation}
where $Relu(\vec{\mathbf{C}})$ represents the remaining capacity of channels.
\modify{If there are remaining capacity of channels, $Relu(\vec{\mathbf{C}}) = \vec{\mathbf{C}}$, otherwise $Relu(\vec{\mathbf{C}}) = 0$.}

{\em Self-service Acceptance Rate}~(SP) means the percentage of customers who 
accepted the switch-to-self-service suggestions;
\begin{equation}
    SP = \frac{\sum \limits_{i=1} \limits^N \vec{1}_{sp}}{N},
    \label{SP-def}
\end{equation}
where $\vec{1}_{sp}$ represents whether the customer accepts the self-service channel.
\modify{$\sum \limits_{i=1} \limits^N \vec{1}_{sp}$ the total number of customers who accept the self-service channels.}

{\em Drainage Acceptance Rate}~(DP) means the percentage of customers who 
accepted the switch-to-app suggestions;
\begin{equation}
    DP = \frac{\sum \limits_{i=1} \limits^N \vec{1}_{dp}}{N},
    \label{DP-def}
\end{equation}
where $\vec{1}_{dp}$ represents whether the customer accepts the target drainage channel.
\modify{$\sum \limits_{i=1} \limits^N \vec{1}_{dp}$ means the total number of customers who accept the drainage channels.}

\modify{Besides, we also calculate the rewards of all RL models for this task, and then normalize these rewards.}

When evaluating on the real dataset, we are additionally interested in 
\textit{Routing Rate}, which means the number of customers who agree to be 
routed to other channels outside hotline, divided by the total number of customers.
These metrics are used to evaluate the reinforcement learning algorithms and traditional machine learning.
\subsection{Customer Flow Forecasting}
\label{subsec-flow-results}
The flow prediction is verified as an important factor to the model in the simulation results listed in the previous section. It is directly related to the performance of the system.

As mentioned in section~\ref{subsec-flow-forecasting}, {\lstm} algorithm~\cite{8632898} and {\xgboost} 
algorithm~\cite{8329419} are widely used in this area, so they are used and compared in our experiments. 

Our pre-processing steps are as follows:
we split the entire flow for every 10 minute and count the flow volume,
this results in a total number of 10{,}271 time points;
In order to predict the volume in the next time window, 
the forecasting makes use of historical data in the past 24 hours (144 points).
All methods are trained on the 80\% of the data and tested on the rest.
 
We follow \cite{DBLP:journals/corr/AzzouniP17} to implement the {\lstm} method, 
which consists of two layers. 
The algorithm runs for 40 epochs with batch size 200.
A dropout rate of 0.8 is employed to avoid overfitting.
The {\xgboost} ends after 40 rounds training, with the learning rate 0.1;
the maximal tree depth is 6; and the colsample-tree is 0.7.
We set an equal weight of 0.1 to both L1 and L2 regularization when
training {\xgboost}.

\begin{figure}[t]
    	\centering
    	\includegraphics[scale=0.185]{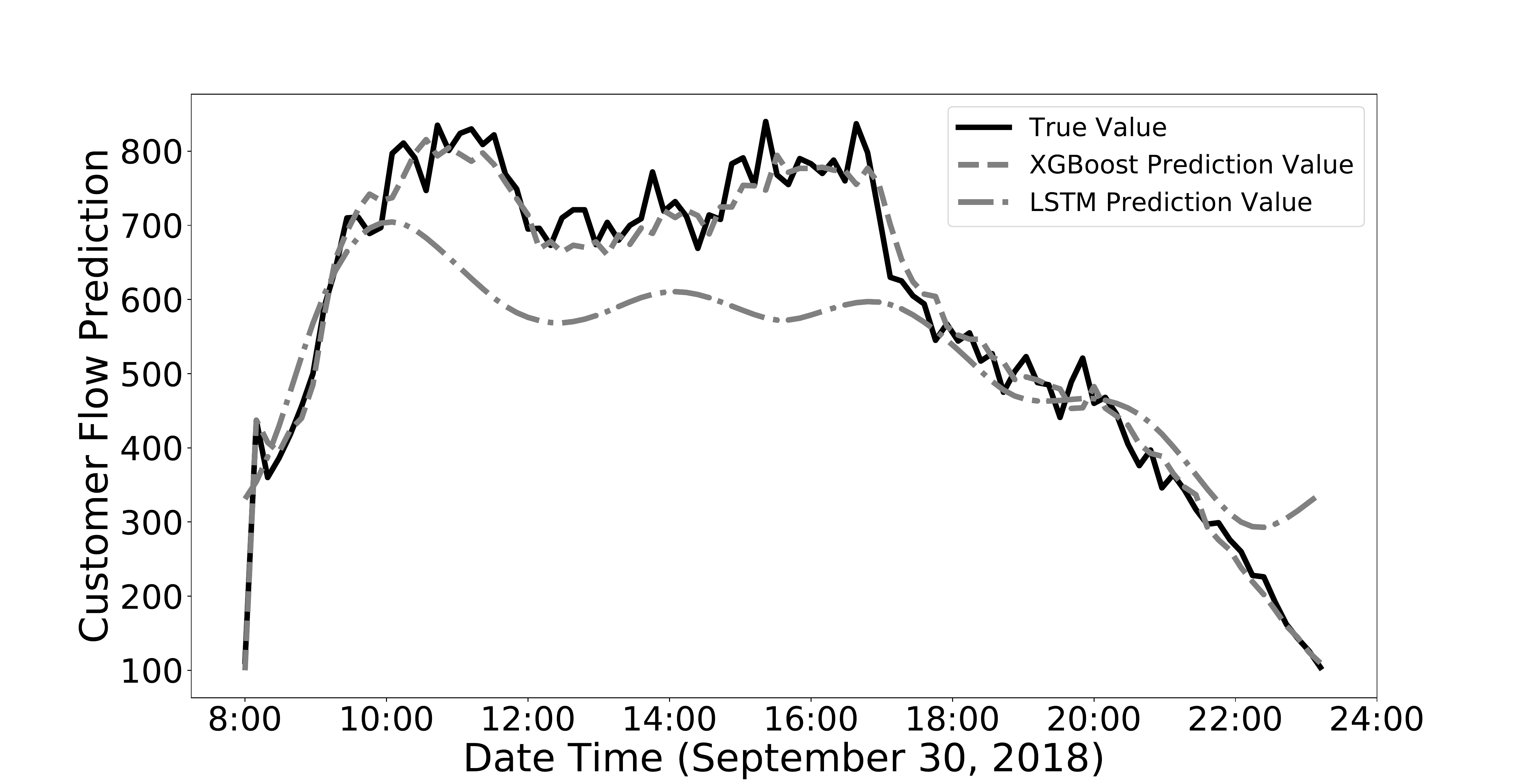}
    	\caption{Comparison of algorithms on the flow forecasting task. The solid 
    	black line is the true value, the dash is result estimated using {\xgboost} and 
    	the dash dot shows the result computed using {\lstm}.}
    	\label{customer-flow}
    \end{figure}

The final prediction results in Figure~\ref{customer-flow}. 
We can observe a better performance of using {\xgboost} than using {\lstm}\cite{DBLP:journals/tvt/MaYCG19}.
For a better understanding of the performance difference, we evaluate both
algorithms using  commonly used metrics, accuracy and
 root-mean-square-error (RMSE):
\begin{equation}\label{RMSE-eq}
\mbox{RMSE} = \sqrt[]{\sum \limits_{i=1} \limits^T \frac{1}{T}(y_i-\hat{y_i})^2},
\end{equation}

where $T$ is the total time points, $y_i$ is the true value and $\hat{y_i}$ is the predict value.
We define the accuracy with a small error-tolerance range: -8\% to 15\%.
For example, when actually there are 100 
customer requests, we regard prediction that falls between 92 and 115 are
correct.

Evaluation results further confirm the observations made in  
Figure~\ref{customer-flow}:
{\xgboost} is more effective, which has an accuracy of {0.803} and
RMSE of {40.445}; while {\lstm} has a lower accuracy and a higher RMSE, which 
are {0.726} and {81.234}, respectively.

\subsection{Routing Results Analysis}
We show the results of various configurations on both real and synthetic
data in this section.

Note that, as in the real application, the bottleneck channel is the
hotline, we will directly use the ``hotline'' channel in results analysis.
\modify{It takes 35.228 seconds training \per-\doddqn's model every 1000 batches and the training convergences after training 4000 batches.}

\begin{figure*}[t]
	\subfigure[Effect of {\per-\doddqn} configurations]{ 
		\label{fig-method-variant} 
		\begin{minipage}[b]{0.5\textwidth} 
			\centering 
			\includegraphics[scale=0.2]{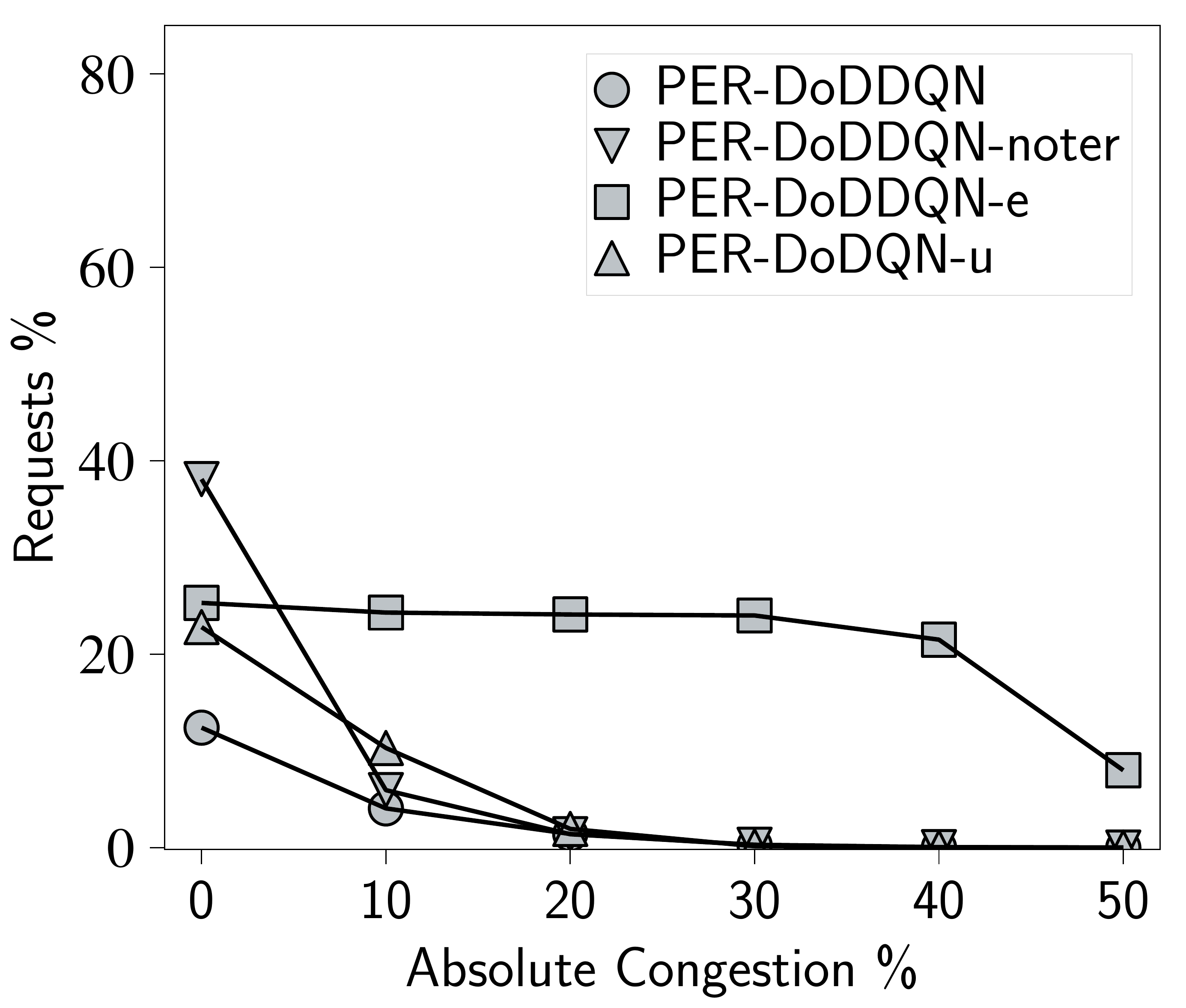} 
	\end{minipage}}%
	\subfigure[Effectiveness comparison of  {\dqn} variants]{ 
		\label{fig-dqn-variant} 
		\begin{minipage}[b]{0.5\textwidth} 
			\centering 
			\includegraphics[scale=0.2]{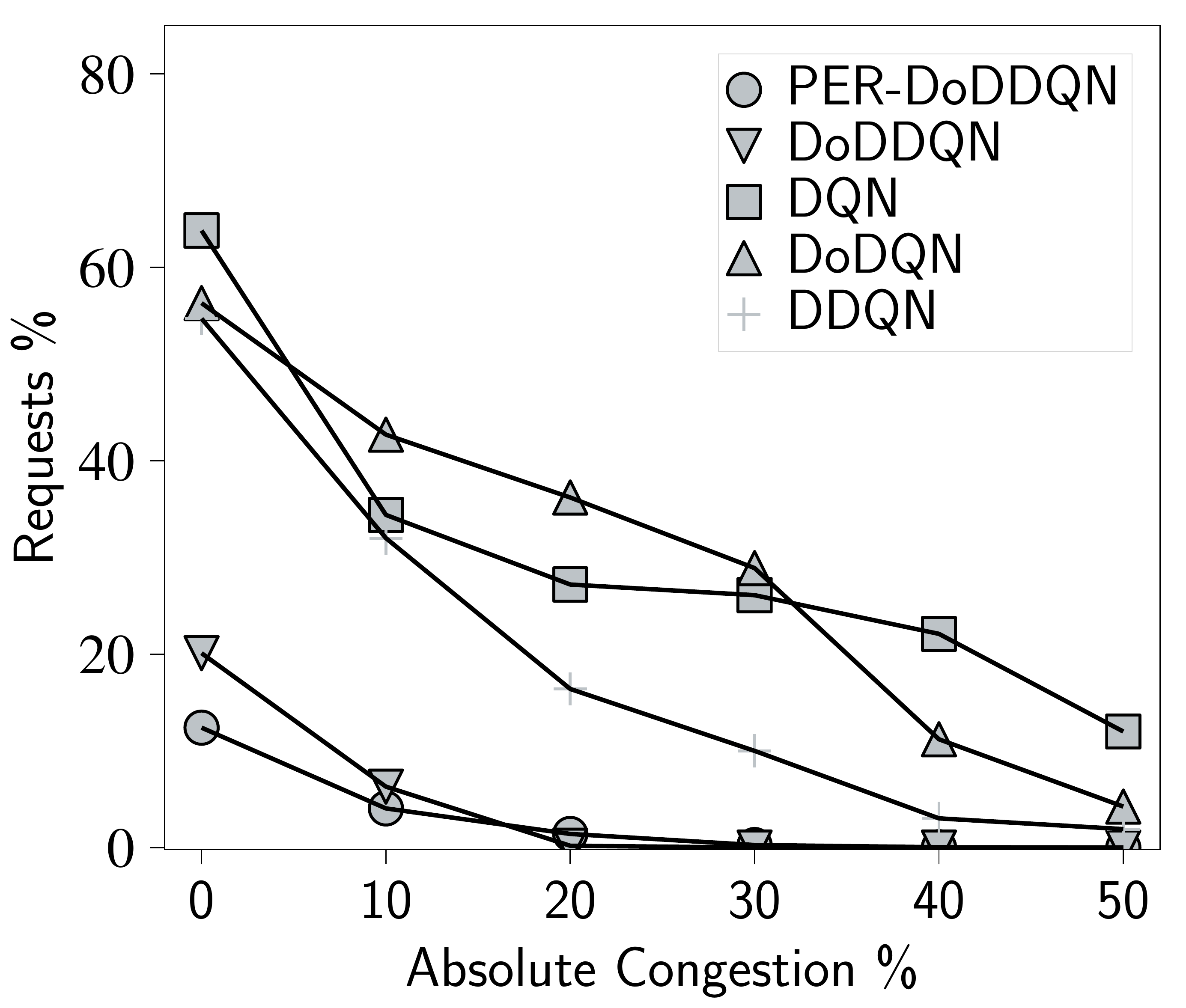} 
	\end{minipage}}%
	\mycaption{Evaluation results on the hotline channel, measured
		by absolute congestion percentage, which is average queue length
		 relative to the channel's original capacity.}
	\label{fig-ac-simulation}
\end{figure*}

\myparagraph{Results on Synthetic Data}
We show evaluation results on the synthetic dataset in 
table~\ref{tbl-syn-result} and figure~\ref{fig-ac-simulation}.
As the hotline channel is the main concern in the real production system,
we first focus on the absolute congestion percentage of this particular
channel, and show results in Figure~\ref{fig-ac-simulation}.
We show the comparison of our proposed {\per-\doddqn} in 
Figure~\ref{fig-method-variant}.
As we can see, our environment model is crucial for finding the 
optimal routing plan: the method {\per-\doddqn-e}, which doesn't
have flow estimation is consistently worse than the others.
Without user modeling also hurt the performance, but only slightly on
the synthetic data.
We then compare our method to other {\dqn} variants in 
Figure~\ref{fig-dqn-variant}.
The trend is clear -- our proposed {\per-\doddqn} is the best
among all.
Also, we can see that the prioritized experience replay plays an
important role in the model, without which the performance can be 
degraded a lot.

\begin{figure}[t]
	\centering 
	\includegraphics[scale=0.2]{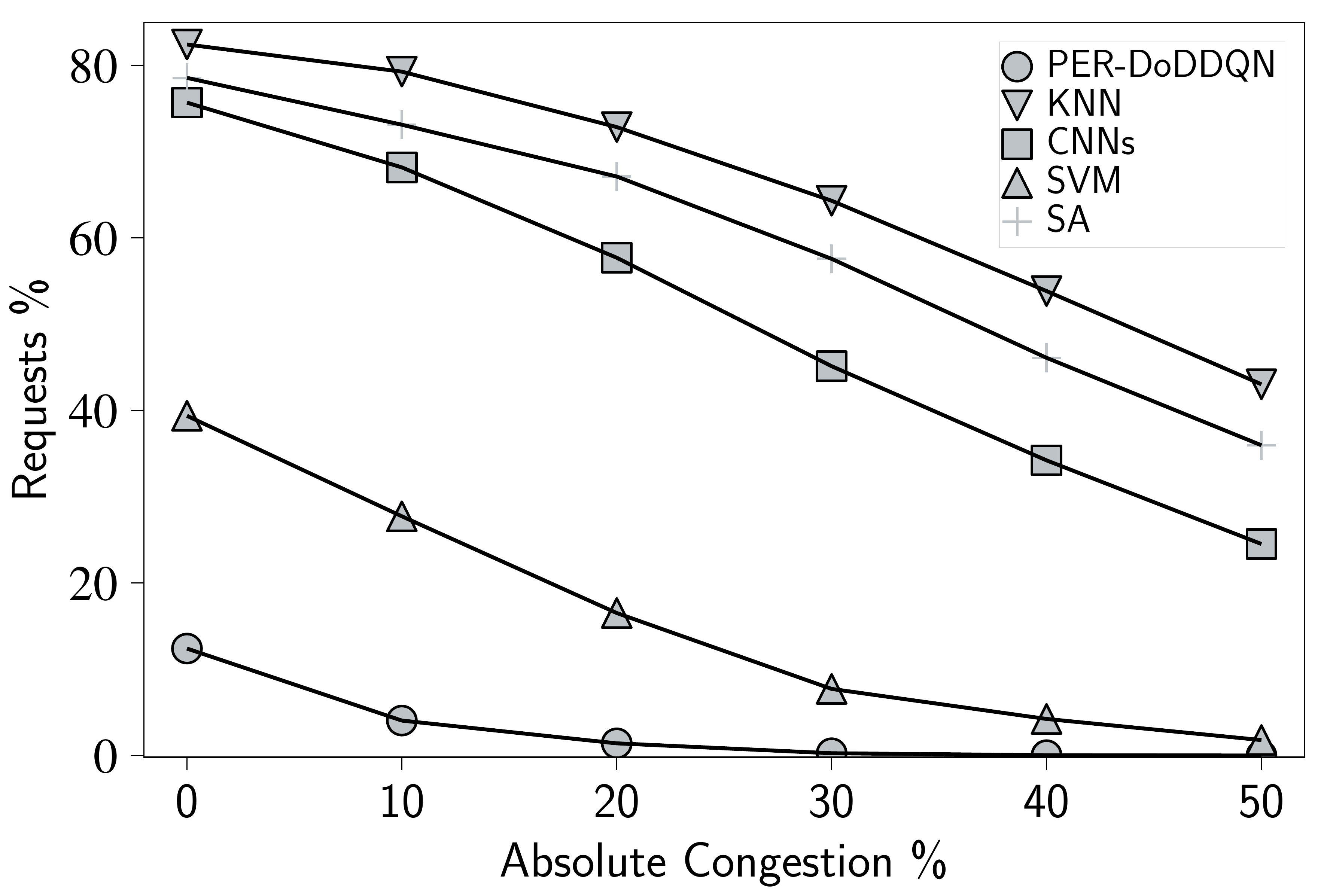} 
	\mycaption{The results of {\pdoddqn} and other machine learning(eg.heuristic algorithm) algorithms on simulated data.}
	\label{fig-method-ml} 
\end{figure}


%

We then consider the overall evaluation using proposed metrics, in Table~\ref{tbl-syn-result}.
Among all evaluation metrics, the CCR is the most important one, and 
the proposed {\per-\doddqn} is the best across all configurations; 
this trend also holds when evaluating using {AC}.
{\doddqn} shows a better performance when considering the peak 
congestion, and it achieves a similar but worse performance than
its {\per} variant.
Using a simple {\dqn} is the best if only drainage percentage or 
average free rate, however, it simply means a non-optimal plan is shown.
Further, results on both {\doddqn} and {\ddqn} suggest that our proposed
{\per-\doddqn} is more advanced than considering any of the components alone.

 \modify{We compare RL models with standard heuristics algorithms, including Simulated annealing (SA), CNN and Baselines(rule-based system) which can be viewed as heuristic methods.}
As shown in Figure~\ref{fig-method-ml} and Table~\ref{tbl-syn-ml-result}, the traditional machine learning (\modify{SA, SVM, KNN, CNN}) perform worse than the RL algorithms, especially in the metrics CCR, AC and PC. Because the long term gains are not considered in these models. The RL framework takes into account not just immediate reward but also the impact of the selected action in the future. But other ML models recommend channels to the customers solely on the current state.

Note that the average idle degree of the catering staff(AFR) can be very low when the customers are more likely to be assigned to hotline regardless of the capacity of channels. And the percentages of customers who accepted the switch-to-self-service suggestions(SP) and switch-to-app suggestions(DP) are independent of the future state. Supervised machine learning methods only care about the current state, so it can perform similarly to or even perform better than the RL models in one or two of the above three metrics, but obviously it can not change the fact that it can not beat RL algorithms among all the metrics in general.

\begin{table*}[t]
	\centering
	\setlength{\tabcolsep}{4pt}
\begin{tabular}{cllllllllllll}
	\toprule
	\multirow{2}{*}{} & \multicolumn{4}{c}{\per-\doddqn} &  & 
	\multirow{2}{*}{\doddqn} &  & \multirow{2}{*}{\dqn} &  & 
	\multirow{2}{*}{\dodqn} 
	&  & \multirow{2}{*}{\ddqn} \\ \cmidrule{2-5}
	& full & e & u & noter &  &  &  &  &  &  &  &  \\ \midrule
	\multicolumn{1}{l}{CCR} & \textbf{0.124} & 0.253 & 0.228 & 0.383 &  & 0.201 &  & 0.638 &  & 0.523 
	&  & 0.547 
	\\
	\multicolumn{1}{l}{AC} & \multicolumn{1}{l}{\textbf{-1.123}} & \multicolumn{1}{l}{-11.693} 
	& \multicolumn{1}{l}{-2.404} & \multicolumn{1}{l}{-2.196} & \multicolumn{1}{l}{} & 
	\multicolumn{1}{l}{-1.648} & \multicolumn{1}{l}{} & \multicolumn{1}{l}{-15.925} & 
	\multicolumn{1}{l}{} & \multicolumn{1}{l}{-15.342} & \multicolumn{1}{l}{} & 
	\multicolumn{1}{l}{-9.179} \\
	\multicolumn{1}{l}{PC} & \multicolumn{1}{l}{-44} & \multicolumn{1}{l}{-67} & 
	\multicolumn{1}{l}{-39} & \multicolumn{1}{l}{-42} & \multicolumn{1}{l}{} & 
	\multicolumn{1}{l}{\textbf{-26}} & \multicolumn{1}{l}{} & \multicolumn{1}{l}{-92} & 
	\multicolumn{1}{l}{} & \multicolumn{1}{l}{-77} & \multicolumn{1}{l}{} & {-77} \\
	\multicolumn{1}{l}{AFR} & 19.606 & 20.955 & 30.775 & 10.383 &  & 33.625 &  & \textbf{8.921} &  & 
	9.360 &  & 
	10.541 \\
	\multicolumn{1}{l}{DP} & 0.258 & 0.325 & 0.245 & 0.307 &  & 0.321 &  & \textbf{0.327} &  & 0.299 
	&  & 0.304 
	\\
	\multicolumn{1}{l}{SP} & 0.400 & 0.373 & 0.395 & 0.392 &  & 0.457 &  & 0.419 &  & 0.385 &  & 
	\textbf{0.478} \\
	\multicolumn{1}{l}{{\modify{Rewards}}} & \textbf{0.912} & 0.465 & 0.649 & 0.767 &  & 0.872 &  & -1.853 &  & -1.334 &  & 
	-0.373

	\\ \bottomrule
\end{tabular}

	\mycaption{
		Synthetic results. The numbers in bold are the best performance on the metric.
	}\label{tbl-syn-result}
\end{table*}

\begin{table*}[t]
    \centering
    \setlength{\tabcolsep}{4pt}
\begin{tabular}{cllllllllllll}
	\toprule
	\multirow{2}{*}{} & \multicolumn{4}{c}{\per-\doddqn} &  & 
	\multirow{2}{*}{KNN} &  & \multirow{2}{*}{CNNs} &  & 
	\multirow{2}{*}{SVM} &  & \multirow{2}{*}{{ \modify{SA}}}\\ \cmidrule{2-5}
	& full & e & u & noter &  &  &  &  &  &  &  &  \\ \midrule
	\multicolumn{1}{l}{CCR} & \textbf{0.124} & 0.253 & 0.228 & 0.383 &  & 0.824 &  & 0.757 &  & 0.394 & & 0.786
	\\
	\multicolumn{1}{l}{AC} & \multicolumn{1}{l}{\textbf{-1.123}} & \multicolumn{1}{l}{-11.693} 
	& \multicolumn{1}{l}{-2.404} & \multicolumn{1}{l}{-2.196} & \multicolumn{1}{l}{} & 
	\multicolumn{1}{l}{-45.085} & \multicolumn{1}{l}{} & \multicolumn{1}{l}{-30.486} & 
	\multicolumn{1}{l}{} & \multicolumn{1}{l}{-7.900} & \multicolumn{1}{l}{} & \multicolumn{1}{l}{-38.773}\\
	\multicolumn{1}{l}{PC} & \multicolumn{1}{l}{-44} & \multicolumn{1}{l}{-67} & 
	\multicolumn{1}{l}{\textbf{-39}} & \multicolumn{1}{l}{-42} & \multicolumn{1}{l}{} & 
	\multicolumn{1}{l}{-161} & \multicolumn{1}{l}{} & \multicolumn{1}{l}{-140} & 
	\multicolumn{1}{l}{} & \multicolumn{1}{l}{-82} & \multicolumn{1}{l}{} & \multicolumn{1}{l}{-148}\\
	\multicolumn{1}{l}{AFR} & 19.606 & 20.955 & 30.775 & 10.383 &  & \textbf{9.937} &  & 12.146 &  & 
	23.104 & & 14.386\\
	\multicolumn{1}{l}{DP} & 0.258 & 0.325 & 0.245 & 0.307 &  & 0.276 &  & \textbf{0.352} &  & 0.273 & & 0.341
	\\
	\multicolumn{1}{l}{SP} & \textbf{0.400} & 0.373 & 0.395 & 0.392 &  & 0.396 &  & 0.392 &  & 0.394 & & 0.386
	\\
	\bottomrule
\end{tabular}

    \mycaption{Synthetic results of {\pdoddqn} and other machine learning algorithms. The numbers in bold are the best performance on the metric.}
    \label{tbl-syn-ml-result}
\end{table*}

\myparagraph{Comparison on the Real Dataset}
We show experimental results on the real dataset, with the real
product system in use as the baseline.
Note that, we use the results of {\xgboost} discussed in
Section~\ref{subsec-flow-results} when modeling environment, in 
all of deep reinforcement learning method.
Among all 4,898,143 records in our dataset, when considering
the baseline model, around 984,526 (20.1\%) of 
customers agree to be routed to self-service, and
only  104,820 (2.14\%) customers are successfully routed to drainage channel.
\begin{table*}[t]
	\centering
	\setlength{\tabcolsep}{3.5pt}
\begin{tabular}{ccccccccccccccccc}
	\toprule
	\multirow{2}{*}{} & \multicolumn{4}{c}{\per-\doddqn} &  & 
	\multirow{2}{*}{\doddqn} &  & \multirow{2}{*}{\dqn} &  & 
	\multirow{2}{*}{\dodqn} 
	&  & \multirow{2}{*}{\ddqn} & & \multirow{2}{*}{Baseline}\\ \cmidrule{2-5}
	& full & e & u & noter &  &  &  &  &  &  &  &  &  & \\ \midrule
    
	DP & 0.341 & 0.282 & 0.277 & 0.341 
		&& 0.340 &  & 0.326 &  & 0.328 && \textbf{0.342} & & 0.021
	\\
	SP & 0.411 & 0.399 & 0.393 & 0.408 
		&& 0.437 &  & \textbf{0.471} &  & 0.449 &  & 0.408 & & 0.201
    \\
  RR
     & \textbf{0.390} & 0.265 & 0.343 & 0.385 
     	&& 0.276 &  & 0.221 &  & 0.263 &  & 0.273 & & 0.216
     \\
  RN
     & \textbf{1.908} &1.297 & 1.680 & 1.889 
     	&& 1.350 && 1.082  && 1.288 && 1.338 && 1.056
     \\
  {{\modify{Rewards}}}
     & \textbf{2.638} &-0.430 & -0.523 & -0.441
     	&& -0.293 && -0.306  && -0.312 && -0.333 && --
	\\ \bottomrule
\end{tabular}

	\caption{Evaluation results on the real dataset.
	The baseline is based on business rules from the product system.
Besides the metrics used previously, the routing rate (RR) and
routing number (RN) is also used.
Note that RN is shown in the unit of $\times 10^6$.}
	\label{Rl-result}
\end{table*}

\begin{table*}[t]
    \centering
    \setlength{\tabcolsep}{3.5pt}
\begin{tabular}{ccccccccccccccccc}
	\toprule
	\multirow{2}{*}{} & \multicolumn{4}{c}{\per-\doddqn} &  & 
	\multirow{2}{*}{KNN} &  & \multirow{2}{*}{CNNs} &  & 
	\multirow{2}{*}{SVM} &  & \multirow{2}{*}{\modify {SA}} 
	\\ \cmidrule{2-5}
	& full & e & u & noter &  &  &  &  &  &  &  &   \\ \midrule
    
	DP & \textbf{0.341} & 0.282 & 0.277 & 0.341 
		&& 0.275 &  & 0.329 &  & 0.290 & & 0.329
	\\
	SP & \textbf{0.411} & 0.399 & 0.393 & 0.408 
		&& 0.395 &  & 0.342 &  & 0.390 & & 0.340
    \\
  RR
     & \textbf{0.390} & 0.265 & 0.343 & 0.385 
     	&& 0.166 &  & 0.109 &  & 0.242 & & 0.119
     \\
  RN
     & \textbf{1.908} &1.297 & 1.680 & 1.889 
     	&& 0.812 && 0.535  && 1.182 & & 0.580
	\\ \bottomrule
\end{tabular}

    \caption{Evaluation results on the real dataset with machine learning algorithms.
	The baseline is based on business rules from the product system.
RR is Routing rate and RN is routing number.
Note that RN is shown in the unit of $\times 10^6$.}
    \label{Ml-result}
\end{table*}

As suggested by Table~\ref{Rl-result}, our proposed  
deep reinforcement learning framework outperforms baseline greatly.
Among all methods {\pdoddqn}, works the best in 
 absolute numbers of customers assigned to non-hotline 
 if there is a  congestion:
 it successfully  reassigns 1.908 million (39.0\%)
  customers to non-hotline channels.
Note that, in real practice, the last tow metric routing rate and
routing numbers are the most important indicators; and
the {\pdoddqn} is the best on both.
Compare to the synthetic data in Table~\ref{tbl-syn-result},
we can see that the {\pdoddqn} is the optimal configuration 
across all methods.
For example, the DP results of {\pdoddqn} is almost the best on the real dataset, while this is not the case on the synthetic dataset.
A similar observation can be made when considering SP metric.
When a user model is built and employed in the framework, we can 
see that it is of great importance in our model -- without which the 
model performance will decrease largely.
The same observation on the importance of flow forecasting can 
also be seen from the effectiveness score of {\pdoddqn-e} in 
Table~\ref{Rl-result}.

Comparing with RL frameworks, the other machine learning algorithms perform poorly in real-world dataset. The number of customers redistributed to other channels by other machine learning algorithms is far less than that by RL models. As shown in Table~\ref{Ml-result}, among all supervised algorithms, SVM performs best in these two metrics (RR and RN). But It only redistributes 1.182 million customers to non-hotline channels, slightly higher than baseline. For DP and SP, supervised algorithms perform worse than {\pdoddqn}.

\section{Related Work}
\label{sec-related-work}
In this section, we will introduce two aspects of technical background in detail, including deep reinforcement learning in Section~\ref{subsec-drl} and flow forecasting in Section~\ref{subsec-flow-forecasting}.

\subsection{Deep Reinforcement Learning}
\label{subsec-drl}
Reinforcement learning (RL) is widely used in many application systems, such as network communication~\cite{DBLP:conf/infocom/XuTMZWLY18}, object detection~\cite{DBLP:conf/iccv/RaoL017}, digital image steganalysis~\cite{8643866} and edge computing~\cite{KDS}.
It is a hot area of machine learning concerned with how 
software agents ought to take actions in an environment so as to maximize some 
notion of cumulative rewards by optimizing the policy~\cite{Sutton+Barto:2018}:
\[\pi_\theta(s,a) =\argmax\limits_\pi 
E(\sum_{t=0}^{\infty}\gamma^t r_t;\theta)
\]
 where $\gamma$ is the 
discounted factor with the immediate reward greater. 
Unlike supervised learning which requires 
labeled training data, or unsupervised one without labels,  
reinforcement learning learns from the environment through ``interaction'': it will 
observe the environment all around, and summarize them into a 'state'($S_t$).  
Together with two strategies of exploration and exploitation, the action $a_t$ 
will be selected. After that action, the agent will observe a new state $S_{t+1}$, and get the rewards $R_t$ of the new environment\cite{DBLP:RL_def}. $R_t$ will be used to update the agent's strategies. After trying a large number of the above steps, the 
agent will optimize its strategies to adapt the environment. Reinforcement learning also defines some 'terminal states'. The learning process will reset a new episode when the agent has reached a terminal state.  Now there are two types of reinforcement: value-based models and policy-based models\cite{Sutton+Barto:2018}.

\label{subsec-qlearning}
Value-based reinforcement learning algorithms aim at learning the state-action value 
function (or Q-function), by minimizing the Temporal-Difference error(often referred to 
as TD-error). The Q value function\cite{Sutton+Barto:2018} can be defined as Equation ~\ref{Q-value-function}:  
\begin{equation}
    q_\theta(s,a) = E(G_t|S_t = s, A_t = a;\theta) 
    =E(r+\gamma E_{a'}(q_\theta(s',a'))),
    \label{Q-value-function}
\end{equation}
 Note that $q_\theta(s,a;\theta)$ is the value function when the agent selects the action $a$ in 
the state $s$, and $s'$ is the new state of the environment. Besides, $\gamma$ is the discounted 
future reward factor, and $r$ is the immediate reward. Value-based algorithms can only deal with enumerable action space. The following four value-based methods are widely used.

\myparagraph{Q-learning}
	The earlier classic RL algorithm is Q-learning\cite{Watkins:1989}, which first 
	generates Q table and R table, then update Q table during training, 
	and therefore it only works in discrete state and action. R table is initialized to 
	the immediate rewards of state-action pairs. While the Q values of 
	the state-action pairs are stored in Q table. Due to Equation \ref{Q-value-function}, the Q 
table is updated during training as follows:
	\begin{equation}
	q(s,a) = q(s,a)+ \alpha(r + \gamma \max \limits_{a'} q(s',a') - q(s,a)).
	\label{eqn-qval}
	\end{equation}

\myparagraph{Deep Q-learning}
As one of the most important branches of machine learning, deep learning has developed rapidly in recent years and has been successfully applied in many areas. \cite{DBLP:journals/tgrs/ChengZH16} use RICNN for object detection insensing images. \cite{DBLP:journals/tcsv/HanZH0RW15} model the background with deep autoencoders for object detection.  \cite{DBLP:journals/ijcv/ZhangHLWL16} propose a unified co-salient framework with two highlights. \cite{DBLP:journals/pami/ZhangMH17} propose SP-MIL framework for co-saliency detection. \cite{manual:Imagetext} propose a fusion algorithm of the image feature and the text feature extracted from two separate networks for image classification. \cite{DBLP:journals/access/SiYM16} analyse DNA methylation data by deep learning.

Deep Learning is able to be coupled with reinforcement learning, which greatly enhances the performance of RL.
One of the successful model is deep Q-learning (DQN), which uses deep learning method to approximate Q table, along with two improvements, experience 
	replay~\cite{DBLP:journals/corr/HosuR16} and two separate neural networks~\cite{DBLP:journals/nature/MnihKSRVBGRFOPB15}. One successful 
	application of DQN is computer games. 
	\cite{DBLP:journals/nature/MnihKSRVBGRFOPB15} mentioned that DQN 
	has already reached the human level in 49 games of Atari 2600 game series.
	DQN is the improvement of Q-learning, minimizing the TD-error:
	\begin{equation}\label{eqn-dqn-loss}
	\mathcal{L} = (r + \gamma \max_{a'}q_{\bar{\theta}}(s', a') - 
		q_{\theta} (s, a))^2,
	\end{equation}
	Where $q_{\bar{\theta}}$ is used to evaluate the target value and $q_ \theta$ is used to 
evaluate the current value, and $q_ \theta$ will be assigned to $q_{\bar{\theta}}$ at regular 
intervals. For simplicity, we refer to ($r + \gamma \max_{a'}q_{\bar{\theta}}(s', a')$) as 
$\hat{y_i}$ 
in the following introduction.
	DQN learns the policy by gradient descent, and the gradient of the loss is written as Equation 
\ref{dqn-loss-gradient}:
	\begin{equation}
		\nabla_\theta \mathcal{L} = E_{s,a,r,s'}((\hat{y_i} - q_{\theta} (s, a))\nabla_\theta q_{\theta} 
(s, a)).
		\label{dqn-loss-gradient}
	\end{equation}
	\myparagraph{Double DQN}
	It has improved DQN through selecting the action before evaluating Q value, 
	which can reduce the chance of overestimations \cite{DBLP:conf/aaai/HasseltGS16}. The loss function (TD-error) of Double DQN is modified 
into Equation \ref{eqn-double-loss}, but the learning method is the same as nature DQN. 
	
\begin{equation}\label{eqn-double-loss}
\mathcal{L} = (r + \gamma q_{\bar{\theta}}(s', 
	\mathop{\arg\max}\limits_{a'}{q_{\theta}}(s',a')) - 
	q_{\theta} (s, a))^2.
\end{equation}
	
	\myparagraph{Dueling DQN}
	The Q value in dueling DQN consists of two parts, value function, $V$, and advantage 
function, $A$. The dueling DQN can learn which states are (or are not) valuable, without having to learn how the action effects the state\cite{DBLP:conf/icml/WangSHHLF16}. And The Q value 
in 
dueling DQN can be written as follows:
	\begin{equation}
		q_{\theta,\alpha,\beta}(s,a) = V_{\theta,\beta}(s) + A_{\theta,\alpha}(s,a) - \frac{1}{|\mathcal{A}|}\sum\limits_{a'}A_{\theta,\alpha}(s,a'),
		\label{dueling-q-value}
	\end{equation}
    Where $\theta$ is the sharing parameter of $V$ and $A$, while $\beta$ and $\alpha$ are the 
private parameters of V and A respectively.
    
\myparagraph{Experience Replay and Prioritized Experience Replay}
The replay buffer is often adopted in the reinforcement learning process, in order to
reduce the correlation of the data.
Experienced replay is adopted by~\cite{DBLP:journals/corr/HosuR16},
it randomly samples transitions from previous training in order to make the data be subject 
to stationary function, and to make the neural network in {\dqn} converge easier.
Randomly sampling transitions from the replay buffer may hurt the performance of the algorithm as 
these transitions are not equally weighted.
A straightforward method is to use a biased sampling method, in which the sampling probability is 
proportional to the TD-error -- transitions with higher TD-errors are more likely to be sampled.
Based on this intuition, \cite{DBLP:journals/corr/SchaulQAS15} proposed prioritized experience 
replay.
Further, in order to reduce  the time spent in sorting these samples, these transitions in prioritized experienced replay are stored using the SumTree data structure~\cite{DBLP:journals/corr/SchaulQAS15}. 
Let $p_j$ be TD-error, the  weight   $w_i$ of the $i$-th transition, $(s_i,a_i,r_i,s'_i)$, is computed using:
\begin{equation}\label{eqn-w-pre} 
w_i = p_i^\alpha / \sum_{j}(p_j^\alpha).
\end{equation} 

\myparagraph{Policy based method}
Policy gradient algorithm is the classical policy based algorithm. Compared to value based algorithm, the policy gradient algorithm directly maximizes the expectations of the state value function. Policy based reinforcement learning can also be applied to continuous action space\cite{Peters+Schaal:2008-motor}.

\myparagraph{Actor-critic algorithm (AC)} 
In the previous RL models, the agent can only be updated after an episode in the policy gradient algorithm.Actor-Critic (AC) algorithm solves this problem by combining policy gradient algorithm with deep Q-learning ingeniously. A3C\cite{DBLP:conf/icml/MnihBMGLHSK16}, DDPG\cite{DBLP:journals/corr/LillicrapHPHETS15}, PPO\cite{DBLP:journals/corr/SchulmanWDRK17} and ACKTR\cite{DBLP:journals/corr/abs-1708-05144} are all the new development of RL based actor-critic framework.
\modify{The policy based methods can be used in continuous action space. DQN algorithms are suitable to deal with discrete action space. Also through the explanation of the above-mentioned content, the RL methods based Actor Critic framework, which have the complex structure, are difficult to converge. Therefore, we prefer to choose DQN to solve our routing task in the next sections, according to the characteristics of the problem.}

\myparagraph{The latest progress}
\modify{\cite{DBLP:journals/corr/MishraRCA17} propose a method based on neural network and combined with reinforcement learning to process the scarce data or the task which changes quickly. \cite{DBLP:journals/corr/abs-1802-09477} illustrates the problem of overestimation exists in the actor-critic framework, and proposes a algorithm based on double DQN to limit overestimation. \cite{DBLP:journals/corr/abs-1801-01290} proposes a algorithm called soft actor-critic based on the maximum entropy framework to improve the convergence and reinforcement learning stability. \cite{DBLP:journals/corr/abs-1802-10031} mainly studied the effect of baselines dependent of state-action, especially in this continuous control tasks.}


\myparagraph{Discussion}
So far, we've   summarized some popular value-based deep reinforcement learning methods.
The {\dqn} proposed by \cite{DBLP:journals/corr/HosuR16} is the first successful integration of the 
deep neural network and reinforcement learning, which  benefits a lot from using experienced replay.
It can work on the continuous state space, which is a result of using the deep neural network as the function approximator.
However, the {\dqn} suffers from overestimations, as pointed out by {\cite{DBLP:conf/aaai/HasseltGS16}}.
To solve the problem, {\cite{DBLP:conf/aaai/HasseltGS16}} proposed double {\dqn} (\dodqn), which 
addresses the problem by modifying the updating formula.
The dueling {\dqn}~(\ddqn) improves by separating Q-value into state value and action advantage, which improves the converging rate~\cite{DBLP:conf/icml/WangSHHLF16}.
In our framework, {\dqn} and all its variants can be applied, and we empirically compare their 
performance in this application scenario.

The policy based methods can be used in continuous action space. DQN algorithms are suitable to deal with discrete action space. Also through the explanation of the above-mentioned content, the RL methods based Actor Critic framework, which have the complex structure, are difficult to converge. Therefore, we prefer to choose DQN to solve our routing task in the next sections, according to the characteristics of the problem.


\subsection{Flow Forecast} 
\label{subsec-flow-forecasting}
In our proposed framework, one of the important parts is to predict the future flow of each channel, 
which we model it as a time-series prediction problem. 
    A time series is a series of data points indexed (or listed or graphed) in time 
    order~\cite{martinetz1993neural}.
    Examples of time series are     heights of ocean tides and counts of sunspots.
     In    our task, customer requests  flow can be viewed as a time series, of which 
     the volume may change temporally.
    The problem of time series prediction is a research topic for many years,  which 
    aims at predicting the value at a given time point.
Time-series prediction problem is often addressed by using machine learning 
models, include  Random Forest~\cite{DBLP:journals/eswa/LinWXZ17}, 
    {\gbrt}~\cite{7514015}, 
    {\xgboost}~\cite{Wang2017Electricity}, and Long Short-Term Memory 
    (\lstm)~\cite{DBLP:journals/corr/AzzouniP17}. 

The Random forest is an ensemble algorithm with 
multiple parallel trees to reduce the time spent in training and 
testing~\cite{DBLP:journals/ml/Breiman01}. 
Whereas, random forest algorithm often suffers from overfitting because of the 
inevitable problem that data may be full of noise. 
The {\gbrt} trained with residual data is a kind of tree-based boosting algorithm~\cite{Friedman:2002:SGB:635939.635941}, so 
it only supports serial execution. 
The improved {\xgboost} greatly improves {\gbrt} and  can  run in parallel with the block structure~\cite{DBLP:journals/corr/ChenG16}.

Deep learning has been demonstrated the success in many application scenarios, and also in the 
time-series prediction.
The superiority of applying  {\lstm}~\cite{DBLP:journals/corr/AzzouniP17} to series prediction can hardly 
be generalized to all types of data.
Cautions need to be made when making decisions on whether to adopt the method in the flow 
forecast problem.
For example, \cite{DBLP:conf/icann/GersES01} use {\lstm} to predict Mackey-Glass 
series and the Santa Fe FIR laser emission series, and observe that {\lstm}
does not seem to consistently be the best on all series, especially on  simpler time series prediction 
tasks, it may be less effective.
\cite{DBLP:conf/icann/GersES01} suggest to use {\lstm} only when simpler traditional methods 
fail.
All previously discussed methods can be applied to our problem, and we will provide empirical
evaluation in Section~\ref{sec-experiments}.

\section{Conclusion and Future Work}
\label{sec-conclusion}
We formulate the classic customer request routing problem into 
an optimization problem by considering both channel resources and
customers' satisfaction.
To address the real problem, we proposed a novel framework, 
which is based on the deep reinforcement learning method.
In addition to the framework, we also propose a new routing method
by combining {\ddqn} and {\dodqn} methods.
Extensive experiments on both real and synthetic data show that our
proposed framework greatly improves the existing system and 
our proposed {\pdoddqn} method is the best configuration.


In the future work, we plan to further improve our method from the following
perspectives:
(i) improve our user profiling by understanding users' description of requests, 
instead of considering attributes alone; 
(ii)  we plan to incorporate real-time features into the proposed {\pdoddqn}
model, for a better model of the environment;
(iii) we also plan to generalize our model to more routing or dispatching related 
problems.

\bibliographystyle{IEEEtran}


\begin{IEEEbiography}[{\includegraphics[width=1in,height=1.25in,clip,keepaspectratio]{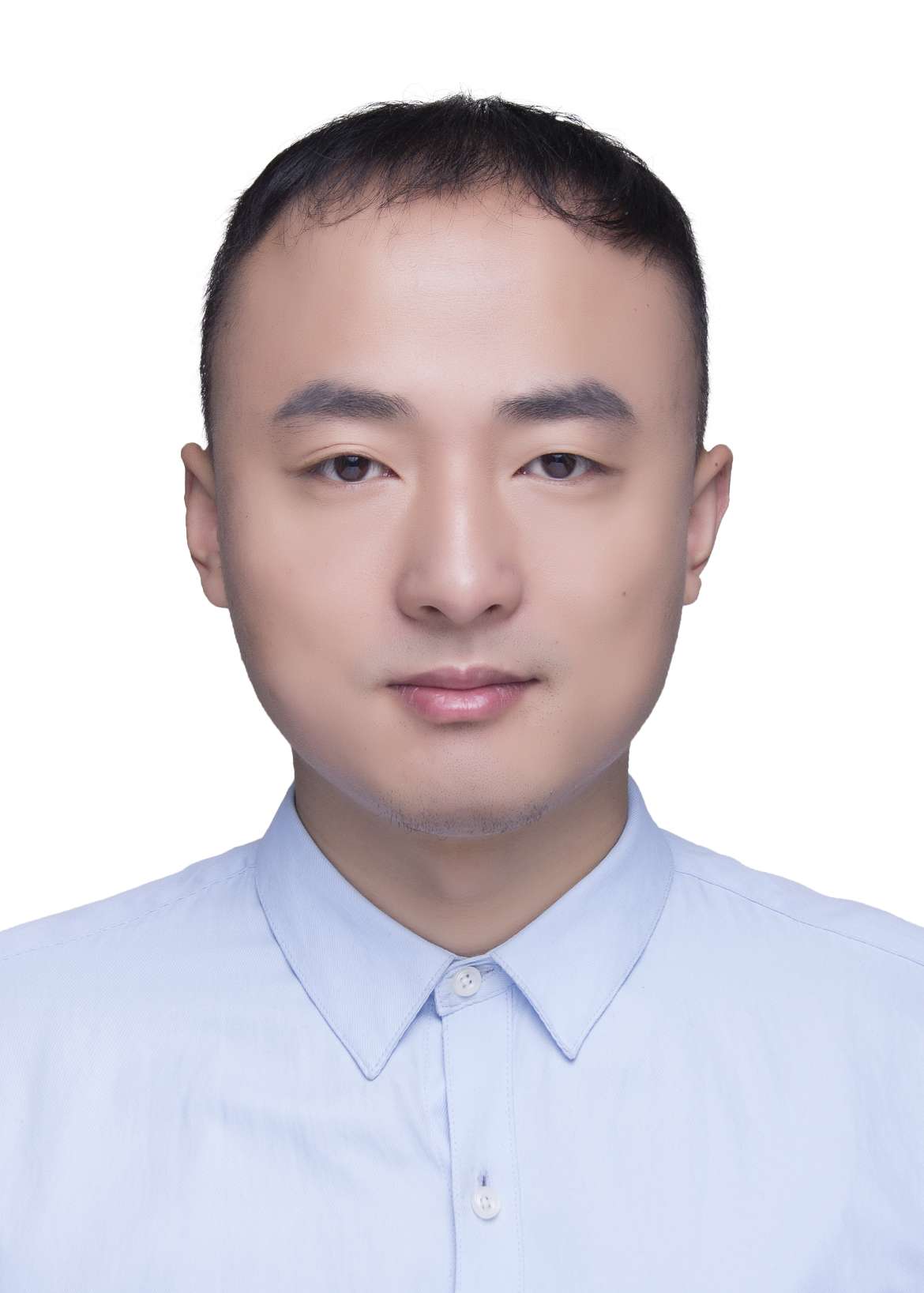}}]{Zining Liu} has been studying for a master's degree at School of Software, Shandong University since September 2016. From July 2017 to July 2018, he worked as an intern at Chinese Academy of Sciences. His research interests include Reinforcement Learning and personalized recommendation system.
\end{IEEEbiography}

\begin{IEEEbiography}[{\includegraphics[width=1in,height=1.25in,clip,keepaspectratio]{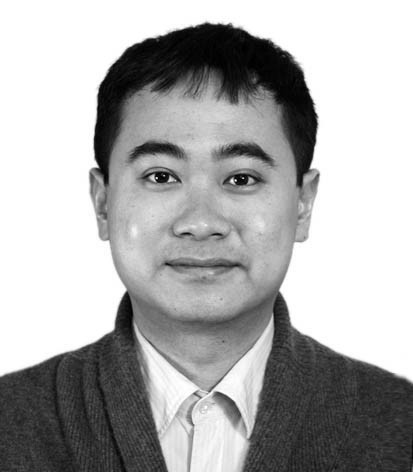}}]{Chong Long} is currently an algorithm expert at Ant Financial Services Group. He received PhD degree from Department of Computer Science and Technology, Tsinghua University, China, in 2010. After that he worked at Yahoo!, Amazon and Hulu, respectively. His research interests include Natural Language Processing, Information Retrieval, Information Extraction, Machine Learning, etc.
\end{IEEEbiography}

\begin{IEEEbiography}[{\includegraphics[width=1in,height=1.25in,clip,keepaspectratio]{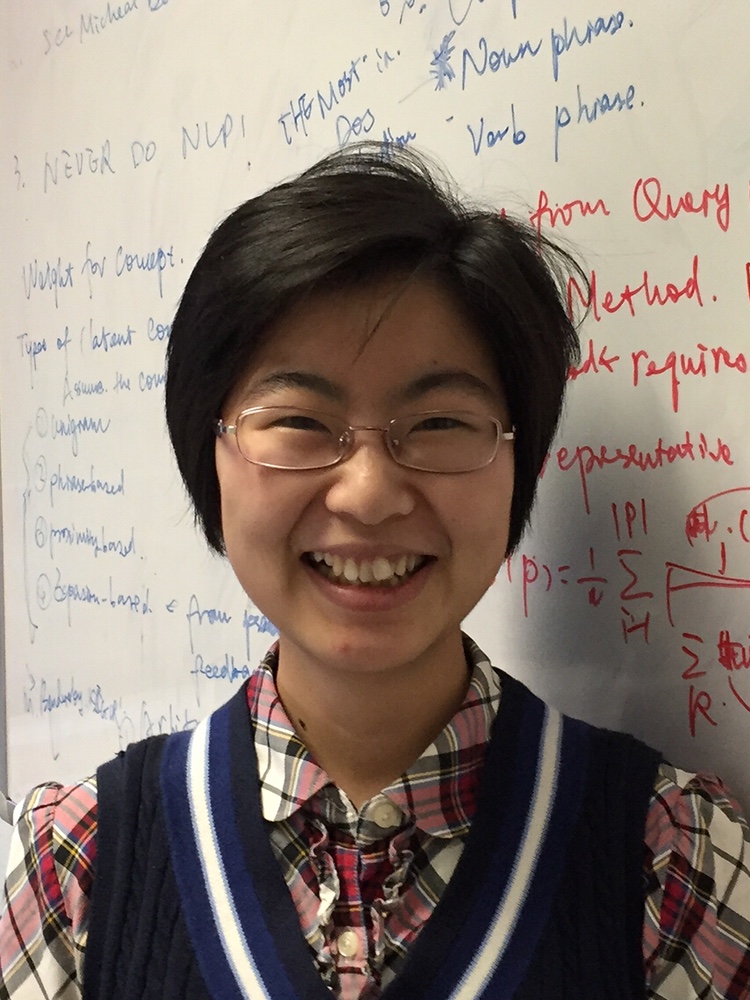}}]{Xiaolu Lu} is a postdoctoral research fellow in Department of Information Technology at RMIT University. She received her PhD in Computer Science from RMIT University in 2018. Her research interest in the area of information retrieval, focusing on search evaluation and trade-offs between effectiveness and efficiency.
\end{IEEEbiography}
\begin{IEEEbiography}[{\includegraphics[width=1in,height=1.25in,clip,keepaspectratio]{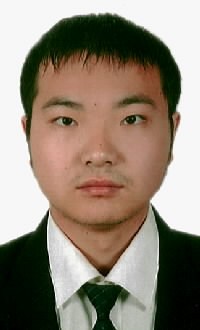}}]{Zehong Hu} is currently an algorithm expert at Alibaba Group. He received Ph.D. from Nanyang Technological University, Singapore. His research interests focus on multiagent systems
and mechanism design. His papers have been published by top conferences, including AAAI, IJCAI and NIPS.
\end{IEEEbiography}
\begin{IEEEbiography}[{\includegraphics[width=1in,height=1.25in,clip,keepaspectratio]{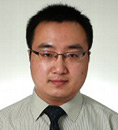}}]{Jie Zhang} is an Associate Professor of the School of Computer Science and Engineering, Nanyang Technological University, Singapore. He is also an Associate of the Singapore Institute of Manufacturing Technology (SIMTech). He obtained Ph.D. in Cheriton School of Computer Science from University of Waterloo, Canada, in 2009. During PhD study, he held the prestigious NSERC Alexander Graham Bell Canada Graduate Scholarship rewarded for top PhD students across Canada. He was also the recipient of the Alumni Gold Medal at the 2009 Convocation Ceremony. The Gold Medal is awarded once a year to honour the top PhD graduate from the University of Waterloo. His papers have been published by top journals and conferences and won several best paper awards. Jie Zhang is also active in serving research communities.
\end{IEEEbiography}

\begin{IEEEbiography}[{\includegraphics[width=1in,height=1.25in,clip,keepaspectratio]{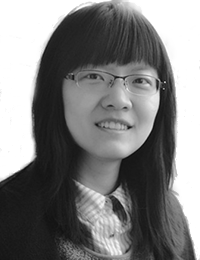}}]{Yafang Wang} obtained her Ph.D. degree in Database and Information Systems group at Max Planck Institute for Informatics, Germany, in February 2013, and worked as a postdoctoral researcher until September 2013. From October 2013 to March 2017, she worked as an associate
professor at the School of Computer Science and Technology, Shandong University. Since the end of March 2017, she has been working as Staff Algorithm Engineer for Ant Financial (Alibaba Group). Her research interests include knowledge harvesting, semantic analytics, reinforcement learning.
\end{IEEEbiography}

\EOD

\end{document}